\pdfoutput=1

\documentclass[11pt]{article}

\usepackage{emnlp2021}
\usepackage{graphicx}
\usepackage{times}
\usepackage{latexsym}
\usepackage{dirtytalk}
\usepackage{enumitem}
\usepackage[T1]{fontenc}

\usepackage[utf8]{inputenc}

\usepackage{microtype}

\usepackage{multirow}
\usepackage{xurl}
\usepackage{csquotes}

\newenvironment{itemize*}%
  {\begin{itemize}%
    \setlength{\itemsep}{0pt}%
    \setlength{\parskip}{0pt}}%
  {\end{itemize}}
  
  \newenvironment{enumerate*}%
  {\begin{itemize}%
    \setlength{\itemsep}{0pt}%
    \setlength{\parskip}{0pt}}%
  {\end{enumerate}}
  
\def\VRdel#1{\bgroup\markoverwith{\textcolor{magenta}{\rule[0.5ex]{2pt}{1pt}}}\ULon{#1}}
\newcommand{\ignore}[1]{}

\def\GAdel#1{\bgroup\markoverwith{\textcolor{blue}{\rule[0.5ex]{2pt}{1pt}}}\ULon{#1}}

\def\ACdel#1{\bgroup\markoverwith{\textcolor{brown}{\rule[0.5ex]{2pt}{1pt}}}\ULon{#1}}

\title{Abuse Detection in Conversational AI} 
\title{Towards Abuse Detection in Conversational AI}
\title{Benchmarking Abuse Detection in Conversational AI}
\title{Abuse towards Conversational AI: Analysis and Detection}
\title{Abusive Language in Conversational AI: Analysis and Detection} 
\title{ConvAbuse: A Dataset for Nuanced Abuse Detection in Conversational AI}
\title{ConvAbuse: Data, Analysis, and Benchmarks for \\ Nuanced Abuse Detection in Conversational AI}

\author{Amanda Cercas Curry$^{1}$ \and Gavin Abercrombie$^{1}$ \and Verena Rieser$^{1,2}$ \\
        $^{1}$The Interaction Lab, Heriot-Watt University, Edinburgh \\
        $^{2}$Alana AI \\ 
        \{ac293, g.abercrombie, v.t.rieser\}@hw.ac.uk}

\begin{document}
\maketitle
\begin{abstract}
\ignore{Abusive language detection has received extensive attention for social media, but considerably less within the context of conversational systems. 
As argued by UNESCO, detecting and mitigating abuse towards these (often anthropomorphised) AI systems is important in order to avoid reinforcement of negative gender stereotypes.
Here, }
We present the first English corpus study on abusive language towards three conversational AI systems gathered `in the wild': an open-domain social bot, a rule-based chatbot, and a task-based system. 
To account for the complexity of the task, we take a more `nuanced' approach where our ConvAI dataset reflects fine-grained notions of abuse, as well as views from multiple expert annotators.
We find that the distribution of abuse is vastly different compared to other commonly used datasets, with more sexually tinted aggression towards the virtual persona of these systems. 
Finally, we report results from bench-marking existing models against this data. Unsurprisingly, we find that there is substantial room for improvement with F1 scores below 90\%.
\end{abstract}

\footnotesize 
\noindent \emph{Warning: This paper contains examples of language that some people may find offensive or upsetting.}
\normalsize

\section{Introduction}

Abusive language detection has received extensive attention for social media, \cite[see e.g.][]{vidgen-etal-2020-detecting}, but far less within the context of conversational systems. 
As argued by UNESCO \cite{west2019d}, detection and mitigation of abuse towards these (often anthropomorphised) AI systems is important in order to avoid reinforcement of negative gender stereotypes.
Following this report, several recent works have investigated possible abuse mitigation strategies \cite{cercas-curry-rieser-2018-metoo, cercas-curry-rieser-2019-crowd,ChinY19,Ma:2019}. However, the results of these studies are non-conclusive as they are not performed with live systems nor with real users -- mainly because of the absence of reliable abuse detection tools. 
The majority of currently deployed systems use simple keyword spotting techniques, \cite[e.g.][]{AmazonAlexaPrize2018,khatri2018detecting}, which tend to produce a high number of false positives, such as cases in which the user expresses frustration, or use of profanities for emphasis, as well as false negatives, e.g.\ missing out on subtler forms of abuse \cite{han-tsvetkov-2020-fortifying}.
Recently, \citet{dinan-etal-2019-build,xu2020recipes} released an abuse detection tool trained on Wikipedia comments and crowd-sourced adversarial user prompts (the latter are not freely available). Whereas in this work, 
\vspace{-0.2cm}
\begin{itemize*}
    \item We show that the distribution of abuse towards conversational systems is vastly different compared to other commonly used datasets, 
   with more than half the instances containing sexism or sexual harassment. 
    \item 
    We develop and release a detailed annotation scheme with the help of experts. 
    \item We use this scheme to annotate a corpus of 20k ratings on $>$6k samples (ca. 2k from each system), 
    which we call ConvAbuse. We critically discuss and experiment with different labelling methods for this task. 
    We also release a subset of 4k examples and their expert annotations.\footnote{We make our code and two of the datasets available at \url{https://github.com/amandacurry/convabuse}. We are unable to release the data from \ignore{open-domain} one system for privacy reasons.}
    \item We benchmark commonly used abuse detection methods on this corpus. 
\end{itemize*}

\section{Related work}\label{sec:related}

Most work on
detecting harmful content such as offensiveness, toxicity, abuse, and hate speech (see \citet{fortuna-etal-2020-toxic} for definitions),
has focused 
on social media platforms, 
foremost  Twitter \citep[e.g.][]{ball-burack-etal-2021-differential,basile-etal-2019-semeval,cao-lee-2020-hategan,davidson-2017-automated,fortuna-etal-2020-toxic,founta2018large,groendahl-2018-all,koufakou-etal-2020-hurtbert,nejadgholi-kiritchenko-2020-cross,nozza-2019-unintended,razo-kubler-2020-investigating,vidgen-etal-2020-recalibrating,wang-etal-2020-detect,waseem-etal-2017-understanding,zampieri-etal-2019-semeval,zampieri-etal-2019-predicting,zampieri-etal-2020-semeval},
or other social media platforms, including
Facebook \citep{glavas-etal-2020-xhate,zampieri-etal-2020-semeval}, Gab \citep{chandra-etal-2020-abuseanalyzer}, and Reddit \citep{han-tsvetkov-2020-fortifying,zampieri-etal-2020-semeval}.
Work has also been undertaken on data from comments on news media \citep{glavas-etal-2020-xhate,razo-kubler-2020-investigating,wang-etal-2020-detect,zampieri-etal-2020-semeval}, chatrooms and discussion forums \citep{gao-etal-2020-offensive,wang-etal-2020-detect}, Wikipedia discussions \citep{fortuna-etal-2020-toxic,glavas-etal-2020-xhate,groendahl-2018-all,nejadgholi-kiritchenko-2020-cross,pavlopoulos-etal-2020-toxicity}, and message services such as WhatsApp \citep{saha2021short}.

Meanwhile, there has been relatively little work
on abuse detection for conversational AI.
Furthermore, much of the work that does exist in this area does not actually involve human-machine dialogue:
\newcite{dinan-etal-2019-build,xu2020recipes} use a classifier developed on Wikipedia comments, which was further trained on adversarial prompts collected via crowd-sourcing.
Similarly, \newcite{delosRiscos2021} designed a chatbot to intervene against online hate speech, trained and evaluated on data from Wikipedia and Civil Comments.

Those few
studies that do report abuse detection results from genuine human-machine conversations tend not to include publicly released datasets.
These include several submissions to the Amazon Alexa Challenge\footnote{\url{https://developer.amazon.com/alexaprize} (accessed May 2021.)} 
\citep{curry2018Alana,khatri2018detecting,paranjape2020neural}.
As such, to the best of our knowledge, this is the first study to release a public dataset of human-machine conversations for
the task in
this domain.

While we aim to detect abuse directed against any target, gender-based abuse has been identified as a particularly prevalent problem in conversational AI \citep{cercas-curry-rieser-2018-metoo,silvervang-2012-effect,west2019d}, and abuse detection systems have themselves been found to contain gender biases \citep{park-etal-2018-reducing}.
Misogyny and sexism detection has been applied to social media in binary \citep{fersini2018overview,nozza-2019-unintended} and multi-class \citep{waseem-hovy-2016-hateful} settings.
We extend this to take an intersectional approach, analysing multiple types of abuse in a hierarchical mutli-label framework (see \S \ref{subsec:guidelines}).

\section{The ConvAbuse corpus} 

We collected data from conversations between users 
and three different conversational AI systems, which have different goals and properties. 
Two of them are classed as chatbots, i.e.\ social, open-domain systems, while the other is a transactional, goal-oriented system.
The first two systems listed below are text-based, whereas the last system is voice-based with a synthetic female-sounding voice. As such, two out of the three systems are female gendered, either by voice or name.

\vspace{-0.2cm}
\paragraph{Alana v2} 
An entrant to the Alexa Challenge 2018, a competition in which university teams develop social chatbots which aim to hold engaging conversations with users in the United States. 
The bot implemented a mixture of social chit-chat and provision of information via entity linking. 
Users were notified of the competition at the beginning of the conversation.
We only have access to the automatically transcribed user utterances, which contain recognition noise. 
The data was collected between April 2017 and November 2018.

\vspace{-0.2cm}
\paragraph{CarbonBot.} 
An assistant created by Rasa\footnote{\url{https://rasa.com} (accessed May 2021.)} and, hosted on Facebook Messenger.\footnote{\url{https://m.me/CarbonBot.from.Rasa} (accessed May 2021.)}
The bot aims to convince the user to buy carbon offsets for their flights. It also notifies the user 
that conversations will be recorded for research purposes.
The data was collected 
between 1st October 2019 and 7th December 2020.

\vspace{-0.2cm}
\paragraph{ELIZA.} An implementation of the rule-based conversational agent intended to simulate a psychotherapist \cite{weizenbaum1966eliza}, designed for academic purposes, and hosted at the Jo\v{z}ef Stefan Institute.
\footnote{\url{http://www-ai.ijs.si/eliza} (accessed May 2021.)} 
It aims to engage the users by asking open questions: \say{\emph{Tell me more about <X>!}}. 
The data was collected between 19th December 2002 and 26th November 2007.

For example conversations from all three systems, see Appendix \ref{app:example_convs}.


\subsection{Pre-processing}

For each system, we discarded any test conversations involving the systems' developers, and extracted the utterances from all user turns from the conversations.
Following the findings of \newcite{pavlopoulos-etal-2020-toxicity} that dialogue context can affect (and even reverse) human judgement of toxicity, we included the system output as well as the previous turns (where available) of both user and system.

We removed any system output that is not directly provided to the user in text form (such as voice prosody tags), and replaced web addresses with the token \texttt{<URL>}.

\subsection{Sampling}
Previous research has shown that 5-30\% of user utterances are abusive \cite{cercas-curry-rieser-2018-metoo}. 
In order to find these instances, one can use purposive nonprobability sampling 
using abusive keywords. 
However, this can lead to the creation of heavily biased datasets \cite{vidgen2020directions,wiegand-etal-2019-detection}.  
We attempted to strike a balance between obtaining a high proportion of examples that contain abusive language and not biasing the datasets towards explicit forms of abuse that contain such keywords. To do this we combined two sets of keywords:
\vspace{-0.3cm}
\begin{enumerate}
    \item {\em A list of `profanities'} --- 265 regular expressions from a blacklist obtained from Amazon. 
    These keywords are mostly profane, offensive words, which can be expected to capture use of explicitly offensive language.  
    \vspace{-0.3cm}
    \item {\em 1,532 terms from Hatebase},\footnote{\url{https://hatebase.org} accessed 6th Nov 2020.} ---  a crowd-sourced list of hate speech  
    to capture (i) abuse targeted at specific groups such as women and racialised minorities, (ii) more subtle forms of abuse that do not contain explicitly offensive language, and (iii) terms that have taken on abusive meanings recently or in certain subcultures.
    As most of the terms also have other, non-hateful meanings \cite{sap-etal-2019-risk}, we hypothesised that their use as keywords could capture abusive content, while not biasing the data towards purely offensive terms.
\end{enumerate}

We then used stratified sampling, to extract utterances at random from six stratas of the datasets that contained conversations featuring 0, 5, 10, 15, 20 and 25 per cent of sentences that feature terms from the list of keywords. 
As the total number of conversations and user turns in CarbonBot is smaller, we did not sample from this, annotating the entire dataset.
We used the bias metrics of \newcite{ousidhoum-etal-2020-comparative}, finding that the final corpus does not  seem to be heavily biased towards typical abusive language keywords (for details, see Appendix \ref{app:sampling}).

\subsection{Annotation scheme and guidelines}\label{subsec:guidelines}

We created a hierarchical labelling scheme based on insights from prior work. 
At the top level, we adapted \newcite{poletto2019annotating}'s unbalanced rating scale, in which input is labelled from $+1$ (\emph{friendly}) to $-3$ (\emph{strongly abusive}), providing information about not only whether or not it is considerd to be abusive, but also the severity of any abuse:

\begin{enumerate}[nolistsep]
\setcounter{enumi}{-4}
    \item Strongly negative with overt incitement to hatred, violence or discrimination, attitude oriented at attacking or demeaning the target.
    \item Negative and insulting/abusive, aggressive attitude.
    \item Negative and impolite, mildly offensive but still conversational.
    \item Ambiguous, unclear. 
    \item Non-abusive.
\end{enumerate}


Based on \newcite{waseem-etal-2017-understanding}'s two-dimensional typology of abuse, we then elicited labels for the \emph{target} (\emph{group}, \emph{individual--system}, or \emph{individual--3rd party}) and \emph{directness} (\emph{explicit} or \emph{implicit}). 
To obtain more finely-grained information about the targets of abuse, annotators then label the instances as either \emph{general}, \emph{sexist}, \emph{sexual harassment}, \emph{homophobic}, \emph{racist}, \emph{transphobic}, \emph{ableist}, or \emph{intellectual}. 
These labels were based on known factors in the matrix of domination \cite{collins2002black}. 
These \emph{type} classes are not mutually exclusive, allowing the annotations to capture intersectionality.  

To allow for contextual interpretations, annotators were shown the target user utterance, the agent's utterance to which it responded, and a previous speaking turn by both the user and the agent.

In supervised learning for text classification tasks, human-provided labels are typically aggregated to one `gold-standard' label per instance by means of majority-vote, adjudication, or statistical methods.
However, the notion of reducing multiple annotations to a single `correct' label has been criticised for erasing minority perspectives \cite{blodgett-2021-socio,gordon2021disagreement}. 
This is because perception of phenomena such as hate, varies both across individuals and culturally \cite{salminen2018online}. 
We therefore retain and evaluate classification systems on the labels of all the annotators. 

\subsection{Annotators}\label{ssc:annotators}

We recruited eight gender studies students in their early 20s. Six of them identify as female, and two as non-binary. 
All are L1 English speakers, predominantly from the United Kingdom, except for one from the United States. 
One identifies as Asian, the remaining seven as white.
Full details are provided in the data statement in Appendix \ref{sec:appendixA}.

\subsection{Agreement measurement and analysis}
We adjusted the annotation scheme iteratively in three rounds by observing the labels applied to batches of 100 random examples from the data.
We measured agreement with Krippendorf's \emph{alpha} ($\alpha$), which can take account for multiple annotators, missing values, and ordinal ratings \cite{gwet2014handbook}. 
Where agreement was low, we invited our experts to discuss examples.
However,  since abuse is a subjective phenomenon, we did not force agreement.
We discarded the data used in guideline development, and the annotators labelled the rest of the data according to the final guidelines.
Agreement scores per annotation task are shown in Table \ref{tab:inter_alpha}. 
Overall, the annotators achieved moderate to substantial agreement for the majority of categories. 
Agreement was consistent across datasets.
We report on intra-annotator agreement in Appendix \ref{app:annotation}.

\begin{table}[ht!]
    \small
    \centering
    \begin{tabular}{l l r}
        \textbf{Annotation task} & \textbf{Label type} &  \textbf{Overall} \\
        \hline 
        Abusive/non-abusive & Binary  &  0.69  \\
        Abuse severity      & Ordinal  & *0.46  \\
        Type                & Binary $\times 8$ &  0.79  \\
        \ \ \ \ Ableism         & Binary  & 0.73  \\
        \ \ \ \ Homophobia          & Binary & 0.83  \\
        \ \ \ \ Intellectual          & Binary &  0.63  \\
        \ \ \ \ Racism          & Binary & 0.96  \\
        \ \ \ \ Sexism          & Binary & 0.63   \\
        \ \ \ \ Sex harassment          &  Binary  & 0.84  \\
        \ \ \ \ Transphobia$\dagger$          & Binary & 0.00  \\
        \ \ \ \ General         & Binary &  0.74 \\
        Target              & Nominal &  0.61  \\
        Directness          & Binary &  0.26      \end{tabular}
    \caption{Inter-annotator agreement: Krippendorf's $\alpha$. *Ordinal weighted $\alpha$ used; $\dagger$ based on only 8 examples.}
    \label{tab:inter_alpha}
    \normalsize
\end{table}

\ignore{
To further validate the labels, we calculate intra-annotator agreement using Cohen's \emph{kappa} ($\kappa$). 
The annotators re-labelled a sample of 10\% of the data, and we calculated \emph{intra}-annotator agreement .
Overall agreement was substantial, but with lower consistency for the \emph{abuse severity} and \emph{directness} labels (see Appendix \ref{app:annotation}).}


\paragraph{Sexism.} Although the annotators form a fairly homogeneous group in terms of demographics and all have a background in Gender Studies, we find only moderate $\alpha$ for sexism, consistent with previous studies which found that up to 85\% of disagreement was on this category \cite{waseem-hovy-2016-hateful}. 
We find that sexism and sexual harassment are closely intertwined but distinct, with 47\% of examples labelled sexist also judged to be sexual harassment but only around 22\% of sexual harassment also being sexist. 
Some annotators see all sexual harassment as necessarily sexist as it is rooted in misogyny. This is in agreement with the European Centre for Gender Equality which states that `sexual harassment is an extreme form of sexism'.\footnote{\url{https://eige.europa.eu/publications/sexism-at-work-handbook/part-1-understand/what-sexual-harassment} (accessed May 2021.)} In our data, sexist examples focus on using 
gendered slurs such as \textit{``bitch''}, 
and sexual harassment uses sex as a way to create a hostile and offensive environment though it may not contain explicit terms, e.g. \textit{``I wanna see you naked''}. 

\paragraph{Directness.} Low inter- but moderate intra-annotator agreement (see Tables \ref{tab:inter_alpha} and \ref{tab:intra}) suggests this task is highly subjective and open to interpretation. 
For example, annotators may perceive abuse as more implicit that is
 phrased as a question (e.g. \textit{``are you stupid'', ``can i be your lover?''}), that is misspelled/misheard, (e.g. \textit{``Connie Lingu''}, 
 or comments with sexual connotations but no overtly sexualised words (e.g. \textit{``call me big daddy''}). 
Annotators can disagree not only on whether abuse is implicit or explicit, but whether it is abuse at all. Examples of disagreement between explicit abuse and non-abuse include commonly used expressions of frustration or surprise such as \textit{``wtf''}. Implicit abuse is particularly difficult to distinguish from non-abusive utterances as annotators must infer the user's tone and intention through capitalisation and punctuation (\textit{``I KNOW!!!!'', ``seems so...''}), or the context 
(\textit{``Does it please you to believe I am stupid? You are a woman, aren't you?''}). 

\subsection{Data and analysis}

We collected a total of 20,710 ratings for 6,837 examples. 
The number of unique examples and labels per dataset is summarised in Table \ref{tab:dataset_summary}. 
Each example is annotated by at least three annotators. 
In order to allow for different points of view to be reflected and modelled, we release the individual ratings in addition to aggregated labels.  
Overall, we find that 27\% of examples have been labelled as abusive (-1 to -3) by at least one annotator, and 20\% of all labels are in this range. 
The subset of examples from Alana v2 have the highest portion of abuse, with 35\% of examples having been labelled as abusive by at least one annotator. 
The target of the overwhelming majority (92\%) of abuse present in our dataset is the system itself.

\begin{table}[ht!] 
    \begin{tabular}{l r r r}
    \textbf{Dataset} & \textbf{Examples} & \textbf{Size} & \textbf{Abuse \%}  \\ 
    \hline
    Total ConvA.        & 6,837 & 20,710 & 20.4 \\
    Alana v2      & 2,652 &  7,942 & 27.2 \\
    CarbonBot  & 1,515 &  4,650 &  6.7 \\
    ELIZA      & 2,670 &  8,118 & 21.2 
\end{tabular}
\caption{Dataset size and labelled examples. Amount of abuse is calculated a total percentage of labels. Note that CarbonBot is not purposively sampled which accounts for the difference.}
\label{tab:dataset_summary}
\end{table}

\begin{figure}[ht!]
 \includegraphics[width=\linewidth]{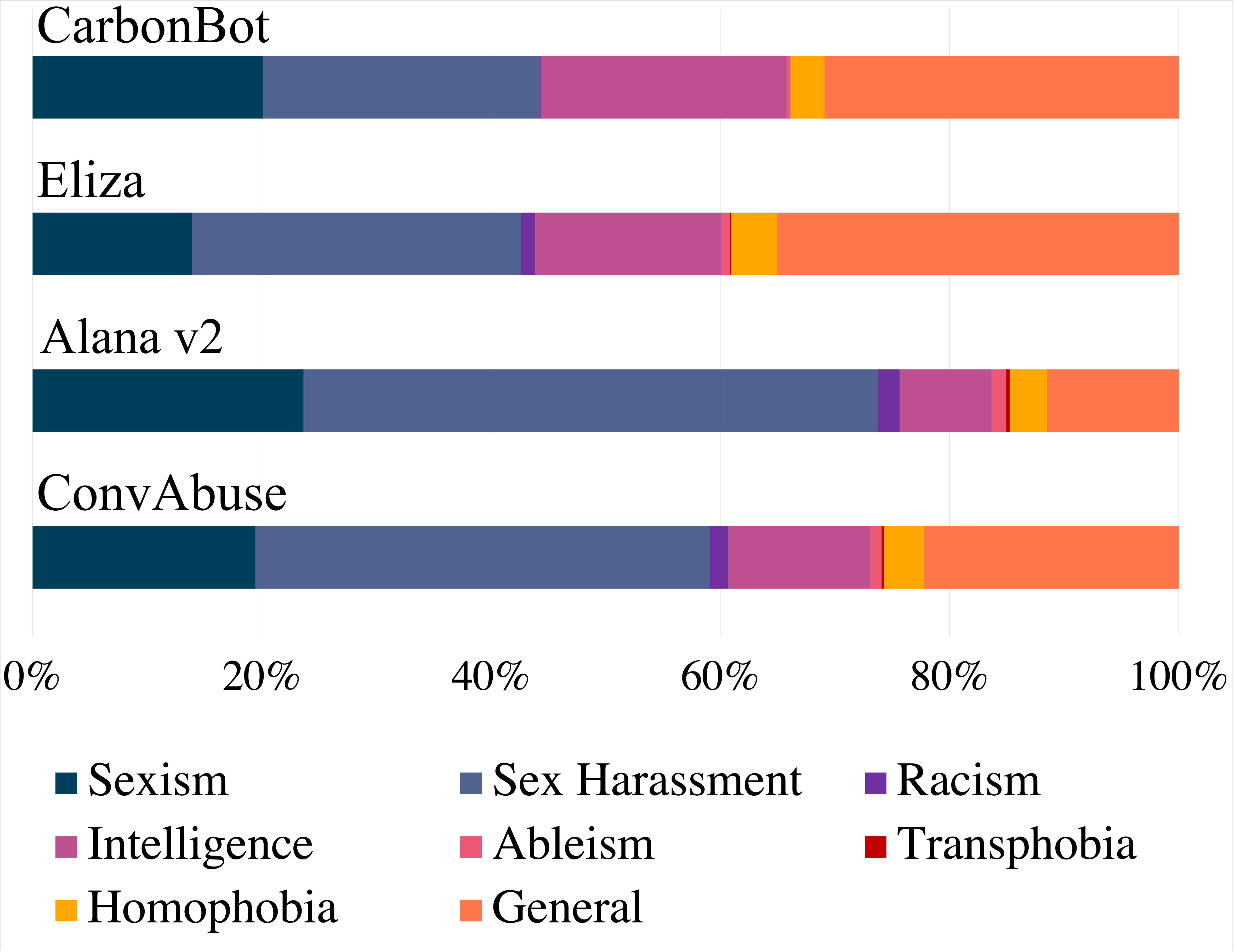}
 \caption{Distribution of abuse types across datasets.}
 \label{fig:abuse-type}
\end{figure}

\paragraph{Abuse type.} Figure \ref{fig:abuse-type} shows the distribution of abuse type labels. Sexual harassment (39.65\%), sexism (19.44\%) and intelligence-based attacks (12.41\%) are the most predominant, while other types are rare at under 5\% of abusive examples (<1\% of total data). 
We attribute this to the personas of the bots, the intimate setting of the interactions, and the gap between the systems' perceived affordances and their actual functionality.
This is supported by the fact that the majority of racism is not directed at the system, but at a third party. 

Although sexual harassment, sexism, and intellectual abuse were  common across systems, Alana v2 (female name and voice) received significantly more sexual harassment and sexist abuse 
than CarbonBot (no gender markers) and ELIZA (female-sounding name), $\chi^2(1, N = 2505) = 67.69, p<0.01$, and $\chi^2(1, N = 3914) = 181.72, p<0.01$, 
respectively. It also received more explicit abuse than the other two systems. 
Conversely, CarbonBot and ELIZA are the target of more intellect-based and `general' abuse. 
This is consistent with previous work showing that female-gendered chatbots receive more sexualised abuse than male ones \cite{brahnam2012gender}, and suggests that the name alone may not elicit strong gender stereotyping. 

\paragraph{Severity.} Severity increases with the number of expletives used 
(\textit{Pearson's r}(20,708)=-.46, p<.001). Similarly, we find that implicit abuse (\textit{``but i think you should quit your job''}) is generally rated as less severe than explicit abuse (\textit{``I think you're an idiot''}): 71\% of implicit abuse is labelled as mildly abusive (-1), whereas only 30\% of explicit abuse is (-1). 
In addition, certain types of abuse are considered more serious that others: 53\% of intellectual abuse is `mild' (-1), compared to 37\% of sexual harassment, 17\% of sexism, and only 7\% of racism which are mainly labelled as `aggressive' (-2) or `attack' (-3). 
See Appendix \ref{app:data_analysis} for more details.

\begin{table*}[ht!]
    \small
    \centering
    \begin{tabular}{p{2.5cm} p{1.8cm}  r  p{2.3cm} r  r  r  c}

         \textbf{Dataset} & \textbf{Source} & \textbf{Size} & \textbf{Labels} & \textbf{Pos. \%} & \textbf{Vocab. size} & \textbf{Utt. length} & \textbf{Context}   \\ 
         \hline \\
         OLID \cite{zampieri-etal-2019-predicting}  & Twitter & 13,240 & offensiveness, targeted, target &33.23 &  9,280 & 24.07 & No\\
         \newcite{ousidhoum-etal-2019-multilingual} & Twitter & 5,647 & abusive, disrespectful, fearful, normal, hateful, offensive & 88.93 &  9,386 &  8.52 & No\\
         Jigsaw Toxic Comment Classification Challenge\footnote{\url{https://www.kaggle.com/c/jigsaw-toxic-comment-classification-challenge}} & Wikipedia Comments &  561,808 & toxic, severe toxic, obscene, threat, insult, identity hate & 6.59 & 157,654 & 64.83 & No \\ \\
         \hline \\
         Alana v2 & Conversational assistant & 2,652 & abuse, type, target, directness & 27.24 & 2,567 & 4.69 & Yes\\
         ELIZA & Chatbot & 2,670 & abuse, type, target, directness & 21.16 & 2,389 & 7.04 & Yes\\
         CarbonBot & Task-oriented chatbot & 1,515 & abuse, type, target, directness & 6.67 & 1,343 & 5.76 & Yes
         
    \end{tabular}
    \caption{Related dataset comparison in terms of data source, dataset size, annotated labels, percentage of positive examples, vocabulary size, utterance length in terms of tokes, and whether there is any interaction context.}
    \label{tab:domain_dist}
    \normalsize
\end{table*}

\subsection{Abuse across domains}

As explored in \S \ref{sec:related}, there has been extensive work in abuse detection and related tasks in social media, particularly Twitter and Wikipedia comments. 
Direct comparison with datasets from other domains is not straightforward as previous studies use different sampling methods, or label slightly different phenomena such as offensiveness and hate speech. 
In this section, we explore how abuse detection in these domains differs by mapping comparable labels across datasets. We do not directly compare the overall proportion of abuse 
but instead describe the datasets in terms of language properties, e.g.\ frequent n-grams, utterance length, vocabulary size, as well as the overall percentage of annotated abuse, see Tables \ref{tab:domain_dist}, and  \ref{tab:common_unigrams} (Appendix \ref{app:data_analysis}).

 \paragraph{Twitter.} 
 While the majority of abuse (92\%) in our dataset is directed towards the system, abuse on Twitter is mainly targeted 
 towards 3rd parties (both individuals and generalised groups). 
 In OLID \cite{zampieri-etal-2019-predicting}, we find that only 46.85\% of abusive instances are in second person (i.e. directed towards the interlocutor), with 36\% and 16\% being directed towards third party groups and individuals, respectively. 
 In terms of (2) directness, we find that the proportions of implicit/explicit abuse are reversed with 89\% of abuse being implicit in \newcite{ousidhoum-etal-2019-multilingual}, compared to only 16\% in our current dataset. 
 Finally, the distribution of abuse types are quite different. 
 Attacks on sexual orientation, disability, and origin are common on Twitter, but are extremely rare in our dataset (<1\% of all labels). In addition, existing Twitter datasets seem to be heavily biased towards explicit language, with similar common words and examples labelled abusive (see Table \ref{tab:common_unigrams} for more details).
 

 \paragraph{Wikipedia comments.} 
Jigsaw's Toxic Comment Classification Challenge\footnote{\url{https://www.kaggle.com/c/jigsaw-toxic-comment-classification-challenge} (accessed May 2021.)} is a competition to identify and classify toxic comments from Wikipedia's talk page edits. 
The data is comprised of over 500k examples labelled in terms of toxicity (see Table \ref{tab:domain_dist}). 
For our analysis, we map the labels `obscene', `threat', `insult', `identity\_hate' to `abusive' based on the definitions given in Jigsaw's Perspective documentation.\footnote{\url{https://developers.perspectiveapi.com/s/about-the-api-attributes-and-languages} (accessed May 2021.)}
Toxic Comments is the largest toxic language dataset and has the largest vocabulary size. 
It's examples are far longer, as they are not limited to a set number of characters, and form part of a discussion. 
In contrast, our ConvAbuse corpus has the shortest utterances, as the systems elicit simpler and more contextual responses. 
In addition, Toxic Comments is heavily biased towards domain-specific language with terms such as `wikipedia', `article' and `edit' among the most common in the dataset.

Overall the source of the data has a significant impact on the language used in the data: while Toxic Comments can be very long, Twitter's character limit clearly impacts the length of the utterances, and 
the utterances in 
ConvAbuse are shorter still and rely more heavily on context. 
 
These varying qualities have implications for the use of such sources as training data for abuse detection tools for conversational systems, such as those developed by \cite{dinan-etal-2019-build} and \cite{xu2020recipes} based on Toxic Comments.
In \S \ref{sec:benchmarking}, we therefore compare the performance of systems with in- and out-of-domain cross-training settings.


\section{Benchmarking} \label{sec:benchmarking}

\ignore{We test a range of standard current machine learning approaches to abuse detection on the ConvAbuse corpus, and experiment with the various in- and out-of-domain combinations 
for cross-training, as well as the inclusion of varying amounts of dialogue context in the inputs.} 

\paragraph{Pre-processing.}

We divide the datasets into train (70\%), validation (15\%), and test (15\%) sets, with similar proportions of positively labelled examples in each split (see Table \ref{tab:train_test_labels}).

\begin{table}[ht!]
    \centering
    \begin{tabular}{l | r | r | r }
         & \textbf{Training} & \textbf{Validation} & \textbf{Testing}  \\
        \hline
        ConvAbuse & 18.31 & 17.77 & 19.29  \\
        Alana v2 & 25.73 & 24.44 &  26.60 \\
        CarbonBot & 5.22 & 4.92 & 6.77  \\
        ELIZA & 18.24 & 18.44 &  19.52
    \end{tabular}
    \caption{Percentage of examples with a positive (abusive) majority label in each split of the data.}
    \label{tab:train_test_labels}
\end{table}

While aggregation of annotators' ratings is problematic (see \S \ref{subsec:guidelines}), it is the dominant paradigm is abusive language detection and NLP in general. 
For comparison, we therefore create a set of aggregated `gold' labels for each (sub-)task based on the majority vote of the annotators on each example.
We evaluate on these in addition to the multiple annotator ratings.
We report the macro-averaged F1 score as an evaluation metric due to the large class imbalances, (e.g.\ most utterances are non-abusive).

\begin{table*}[ht!]
    \centering
    \begin{tabular}{l|rrrrrr}
         &  Agg. & Multi & Agg. & Multi & Agg. & Multi \\
        \textbf{Training data}  & \multicolumn{2}{c}{\textbf{Random class}} & \multicolumn{2}{c}{\textbf{Keywords}} &  \multicolumn{2}{c}{\textbf{Perspective}} \\
         \hline  
        \ \ \ \ \ ---  & 32.07 & 29.68 & 54.56 & 31.70 & 74.04 & 71.66 \\ 
        \hline 
         & \multicolumn{2}{c}{\textbf{SVM}} & \multicolumn{2}{c}{\textbf{MLP}} & \multicolumn{2}{c}{\textbf{BERT}} \\
        \hline 
                                
        ConvAbuse             & 84.35 & 82.68 & 83.84 & 83.58 & 88.92 & 87.86 \\
        Alana v2             & 82.48 & 80.41 & 81.25 & 81.78 & 86.60 & 82.53 \\
        CarbonBot             & 71.98 & 65.66 & 70.63 & 66.98 & 43.38 & 48.07 \\
        ELIZA                 & 82.63 & 76.79 & 81.19 & 79.53 & 87.46 & 83.71 \\
        OLID                  & 32.01 & 30.30 & 31.85 & 27.85 & 37.55 & 23.62 \\
        Toxic Comments large  & 87.22 & 85.73 & 84.71 & 83.59 & 88.41 & 86.47 \\
        Toxic Comments small  & 78.26 & 78.56 & 74.91 & 72.70 & 83.70 & 83.98 
        
    \end{tabular}
    \caption{Macro F1 scores for the binary abuse detection task using aggregated and multiple annotator labels.}
    \label{tab:results_cross}
\end{table*}

\subsection{Models} 

We test the following approaches on the main binary abuse detection task in both the aggregated and multi-annotation settings.
We also assess the performance of the best performing approaches with varying amounts of context, and test a simple neural method on the four sub-tasks.

\subsubsection*{Baselines}

\begin{itemize*}
    \item {\bf Random classifier}: Outputs predicted labels uniformly at random.
    \item {\bf Keyword filtering}: We use the same keyword list as for sampling.
    \item {\bf Jigsaw Perspective}: We test a commercial off-the-shelf system, that has been trained on data including comments on Wikipedia and news articles.\footnote{\url{https://perspectiveapi.com/} (accessed May 2021.)} 

\end{itemize*}

\subsubsection*{Machine learning methods}

Following initial hyperparameter optimization experiments, we use the following systems and settings:

\begin{itemize*}
    \item {\bf Support Vector Machine}: 
    SVMs have been used is previous work on abuse detection in Twitter data, \cite[e.g.][]{davidson-2017-automated}, and have been shown to outperform neural systems \cite{niemann-etal-2020-abusive}.
    We train a linear SVM on bag-of-words representations of the texts using term frequency-inverse document frequency (tf-idf) scores for unigram feature selection. 
    We use \emph{l2} normalisation and set \emph{C}=1.
    \item {\bf Multi-Layer Perceptron}: A standard neural network with one hidden layer consisting of 256 units, ReLu activation, a dropout rate of 0.75, and Adam optimisation with a learning rate of $1e-3$. 
    We use early-stopping to find the best performing model on the validation set.
    We use the same text features as for the SVM.
    \item {\bf BERT}: 
    To account for data sparsity, we use pre-trained BERT embeddings \cite{devlin-etal-2019-bert} and fine-tune the model for four epoques, using a single classification layer and the same architecture as in \cite{dinan-etal-2019-build}.
    We set the learning rate to $1e-4$.
\end{itemize*}


\subsection{Cross-training} \label{sec:cross_training}

To observe the effects of domain shift, we evaluate the systems with different combinations of data from the following sources for training and testing:
\begin{itemize}
    \item Training: OLID \cite{zampieri-etal-2019-predicting}, Wikipedia Toxic Comments (as used by \citet{dinan-etal-2019-build}), Alana v2, CarbonBot, ELIZA, ConvAbuse (all three conversational AI datasets, both individually and combined).
    \vspace{-0.3cm}
    \item Testing: The ConvAbuse corpus, and the subsets Alana v2, CarbonBot, and ELIZA.
\end{itemize}

\begin{table*}[ht!]
    \centering
    \begin{tabular}{c|rrrrrrrr}
           & \multicolumn{2}{c}{\textbf{Severity}} & \multicolumn{2}{c}{\textbf{Type}} & \multicolumn{2}{c}{\textbf{Target}} & \multicolumn{2}{c}{\textbf{Directness}} \\
            & Agg. & Multi & Agg. & Multi & Agg. & Multi & Agg. & Multi \\
        \hline 
        Random & 16.61 & 17.64 & 35.68 & 26.63 & 22.29 & 21.69 & 58.41 & 65.66   \\
        MLP    & 44.86 & 55.43 & 77.56 & 73.50 & 31.71 & 31.72 & 72.30 & 70.41 \\
        BERT   & 50.15 & 54.28 & 73.35 & 49.54 & 32.90 & 31.72 & 77.81 & 70.66
    \end{tabular}
    \caption{Sub-task macro-averaged F1 scores evaluated against the random classifier baseline on the aggregated and multi-annotator labels.}
    \label{tab:results_subtasks}
    \normalsize
\end{table*}

\paragraph{Results} are presented in Table \ref{tab:results_cross}.
We find that the best performance in most training settings is obtained using BERT.
The highest F1 scores are obtained when training in-domain on the ConvAbuse data, or on Toxic Comments (TCs).
However, this dataset is around 40 times larger than any of the other training sets.
When TCs is reduced to a comparable size, the F1 score drops to considerably below that of the ConvAbuse-trained systems. These results
highlight the differences between the two domains and the benefits of training on conversational AI data.
Training on OLID, which is both small 
and out-of-domain, results in the lowest scores. 

\subsubsection{Contextual input features}


The majority of previous work on abuse detection does not take context into account, or provides inconclusive evidence of its importance. 
\newcite{menini2021abuse} showed that the more context is available, the likelier tweets are to be considered non-abusive by annotators.
And \newcite{dinan-etal-2019-build} showed that context improves detection performance (providing six total turns with five of context).
However, \newcite{pavlopoulos-etal-2020-toxicity} found very few examples of toxicity to be context-sensitive for Wikipedia comments, and that inclusion of dialogue context did not lead to large performance gains.

We train and test {the classifiers} on ConvAbuse with: (1) no context (single utterance), (2) the agent's turn (two total turns), and (3) the agent's turn plus the previous turn of both user and agent (four turns).
We concatenate the turns in the inputs
in each setting.

\begin{table}[ht!]
    \centering
    \resizebox{7.7cm}{!}{
    \begin{tabular}{l|rrrrrr}
        & \multicolumn{2}{c}{\textbf{No context}} & \multicolumn{2}{c}{\textbf{+ agent turn}} & \multicolumn{2}{c}{\textbf{+ prev. turns}} \\
        & Agg. & Multi & Agg. & Multi & Agg. & Multi \\
        \hline 
    SVM    & 84.35 & 82.68 & 82.71 & 77.38 & 74.46 & 73.33 \\
    MLP    & 83.84 & 83.58 & 80.09 & 78.37 & 73.99 & 72.83 \\
    BERT   & 88.92 & 87.86 & 88.96 & 86.55 & 88.18 & 84.53
    \end{tabular}}
    \caption{Macro-averaged F1 scores for binary abuse classification with varying amounts of context.}
    \label{tab:results_context}
\end{table}

Results are shown in Table \ref{tab:results_context}.
As more context is added, the performance of both the SVM and MLP degrades, possibly as a result of increased data sparsity.  
However, performance using BERT is similar in all three settings, suggesting that it may be able to better handle the long-range contextual dependencies. 
We leave exploration of more complex classification frameworks that may be able to exploit the contextual information
for future work.

\subsection{Fine-grained abuse detection}

We also provide benchmarks for the four sub-tasks: \emph{severity} (ordinal classification), \emph{type} (multiclass, multilabel classification of the eight categories described in Section \ref{subsec:guidelines}), \emph{target} (ternary) and \emph{directness} (binary).
Here, we use the two neural systems, as they can more easily handle the ordinal labels.
We train on the ConvAbuse dataset, which is labelled for these tasks.

Results are shown in Table \ref{tab:results_subtasks}.
We find that the systems comfortably beat the random baselines for each task, with little difference between the two classifiers.
They both perform poorly on multiple nominal (\emph{target}) and ordinal (\emph{severity}) classes and in some of the multi-annotator settings, which suffer from label sparsity in some of the classes.

We model each abuse \emph{type} as a binary classification task, rather than multi-label prediction, enabling multiple types to be assigned to each example.
We find that the classifier often confuses the  classes \emph{sexism} and \emph{sexual harassment}.
In around half of cases in which the true label is one of these, the system predicts the other.
We leave more focused approaches, like multi-task learning, for future work.

\ignore{
\begin{table*}[ht!]
    \centering
    \small
    \begin{tabular}{ll|rrrrrrrrrrrr}
        \textbf{Test} & \textbf{Train} & \multicolumn{2}{c}{\textbf{Majority}} & \multicolumn{2}{c}{\textbf{Keywords}} & \multicolumn{2}{c}{\textbf{Jigsaw}} & \multicolumn{2}{c}{\textbf{SVM}} & \multicolumn{2}{c}{\textbf{MLP}} & \multicolumn{2}{c}{\textbf{BERT}}\\
        \textbf{data} & \textbf{data} & Agg. & Multi & Agg. & Multi & Agg. & Multi & Agg. & Multi & Agg. & Multi & Agg. & Multi \\
        \hline 
        --- & ConvAbuse       &  & &  & &  & & --- & --- & --- & --- & --- & --- \\
        --- & CarbonBot       &  & &  & &  & & & & &  \\
        --- & ELIZA           &  & &  & &  & & & & &  \\
        --- & Alana v2       &  & &  & &  & & & & &  \\
        \hline 
        --- & ConvAbuse       &  & &  & &  & & & & &  \\
        --- & CarbonBot       &  & &  & &  & & & & &  \\
        --- & ELIZA           &  & &  & &  & & & & &  \\
        --- & Alana v2       &  & &  & &  & & & & &  \\
        \hline 
        --- & ConvAbuse       &  & &  & &  & & & & &  \\
        --- & CarbonBot       &  & &  & &  & & & & &  \\
        --- & ELIZA           &  & &  & &  & & & & &  \\
        --- & Alana v2       &  & &  & &  & & & & &  \\
        \hline
        \hline
        \multirow{6}{*}{ConvAbuse} & ConvAbuse & --- & --- & --- & --- &  & & & & &  \\
                                   & CarbonBot & & &  & &  & & & & &  \\
                                   & Eliza &  & &  & &  & & & & &  \\
                                   & Alana v2 & & &  & &  & & & & &  \\
                                   & Twitter & & &  & &  & & & & &  \\
                                   & Wikipedia & & &  & &  & & & & &  \\
        \multirow{6}{*}{CarbonBot} & ConvAbuse &  & &  & &  & & & & &  \\
                                   & CarbonBot & & &  & &  & & & & &  \\
                                   & Eliza &  & &  & &  & & & & &  \\
                                   & Alana v2 & & &  & &  & & & & &  \\
                                   & Twitter & & &  & &  & & & & &  \\
                                   & Wikipedia & & &  & &  & & & & &  \\
        \multirow{6}{*}{ELIZA}     & ConvAbuse &  & &  & &  & & & & &  \\
                                   & CarbonBot & & &  & &  & & & & &  \\
                                   & Eliza &  & &  & &  & & & & &  \\
                                   & Alana v2 & & &  & &  & & & & &  \\
                                   & Twitter & & &  & &  & & & & &  \\
                                   & Wikipedia & & &  & &  & & & & &  \\
        \multirow{6}{*}{Alana v2} & ConvAbuse &  & &  & &  & & & & &  \\
                                   & CarbonBot & & &  & &  & & & & &  \\
                                   & Eliza &  & &  & &  & & & & &  \\
                                   & Alana v2 & & &  & &  & & & & &  \\
                                   & Twitter & & &  & &  & & & & &  \\
                                   & Wikipedia & & &  & &  & & & & &  
         
    \end{tabular}
    \caption{F1 scores}
    \label{tab:results_main}
    \normalsize 
\end{table*}
}

\ignore{
\begin{table}[ht!]
    \centering
    \small
    \begin{tabular}{l|rrrrrr}
        \textbf{Test} & \multicolumn{2}{c}{\textbf{Majority}} & \multicolumn{2}{c}{\textbf{Keywords}} & \multicolumn{2}{c}{\textbf{Jigsaw}} \\
        \textbf{data} & Agg. & Multi & Agg. & Multi & Agg. & Multi  \\
        \hline 
         ConvAbuse       &  &  &  &  &  &  \\
         CarbonBot       &  &  &  &  &  &  \\
         ELIZA           &  &  &  &  &  &  \\
         Alana v2       &  &  &  &  &  &  \\
        \hline 
         ConvAbuse       &  &  &  &  &  &  \\
         CarbonBot       &  &  &  &  &  &  \\
         ELIZA           &  &  &  &  &  &  \\
         Alana v2       &  &  &  &  &  &  \\
        \hline 
         ConvAbuse       &  &  &  &  &  &  \\
         CarbonBot       &  &  &  &  &  &  \\
         ELIZA           &  &  &  &  &  &  \\
         Alana v2       &  &  &  &  &  &  

    \end{tabular}
    \caption{Baselines F1 scores}
    \label{tab:results_baselines_main}
    \normalsize 
\end{table}
}

\ignore{
\begin{table*}[ht!]
    \centering
    \resizebox{\textwidth}{!}{
    \begin{tabular}{l|rrrrrrrrrrrr}
         \textbf{Training} & \multicolumn{2}{c}{\textbf{Majority}} & \multicolumn{2}{c}{\textbf{Keywords}} &  \multicolumn{2}{c}{\textbf{Jigsaw}} &  \multicolumn{2}{c}{\textbf{SVM}} & \multicolumn{2}{c}{\textbf{MLP}} & \multicolumn{2}{c}{\textbf{BERT}} \\
         \textbf{data} & Agg. & Multi & Agg. & Multi  & Agg. & Multi & Agg. & Multi & Agg. & Multi & Agg. & Multi \\
        \hline 
                                    \ \ \ \ \ \ \ --- &  &  &  &  &  &  & ---\ & ---\ & ---\ \ & ---\ \ & ---\ \ & ---\ \  \\
                                    ConvAbuse         & 32.35\ & ---\ & 54.56\ & ---\ & ---\ & ---\ & 78.08 &  & 72.32 &  73.30 & 67.86 & 65.27 \\
                             Alana v2         & 42.02\ & ---\ & 56.65\ & ---\ & ---\ & ---\ & 80.58 &  & 68.75 & 68.79 & 56.25 & 59.13 \\
                                    CarbonBot         & 12.68\ & ---\ & 45.00\ & ---\ & ---\ & ---\ & 76.92 &  & 52.14 & 45.27 & 51.83 & 49.43 \\
                                Eliza             & 32.66\ & ---\ & 54.20\ & ---\ & ---\ & ---\ & 74.87 &  & 68.99 & 63.64 & 58.38 & 59.77 \\
                                    Twitter           & ---\ & ---\ & ---\ & ---\ & ---\ & ---\ & ---\ & ---\ & 42.22 & ---\ & 34.55 & ---\ \  \\
                                    Wikipedia         & ---\ & ---\ & ---\ & ---\ & ---\ & ---\ & 49.89 & ---\ & 78.79 & ---\ &  & ---\ \ 
        
    \end{tabular}}
    \caption{F1 scores for the binary abuse detection task}
    \normalsize 
\end{table*}}

\ignore{ 
\begin{table*}[ht!]
    \small
    \centering
    \begin{tabular}{lll|rrrrrrrrrr}
        \textbf{System} & \textbf{Train} & \textbf{Test} & \multicolumn{2}{c}{\textbf{Abuse}} & \multicolumn{2}{c}{\textbf{Type}} & \multicolumn{2}{c}{\textbf{Severity}} & \multicolumn{2}{c}{\textbf{Target}} & \multicolumn{2}{c}{\textbf{Directness}} \\
         & \textbf{data} & \textbf{data} & Gold & Multi & Gold & Multi & Gold & Multi & Gold & Multi & Gold & Multi \\
        \hline 
        \multirow{4}{*}{Majority} & -- & ConvAbuse &  &  &  &  &  &  &  &   &  & \\
                                  &    & CarbonBot &  &  &  &  &  &  &  &   &  & \\
                                  &    & ELIZA     &  &  &  &  &  &  &  &   &  & \\
                                  &    & Alana v2     &  &  &  &  &  &  &  &   &  & \\
        \hline 
        Keywords &  & ConvAbuse &  &  &  &  &  &  &  &   &  & \\
                                  &    & CarbonBot &  &  &  &  &  &  &  &   &  & \\
                                  &    & ELIZA     &  &  &  &  &  &  &  &   &  & \\
                                  &    & Alana v2     &  &  &  &  &  &  &  &   &  & \\
        Perspective &  &  &  &  &  &  &  &  &  &   &  & \\
        MLP                   & ConvAbuse & ConvAbuse &  &  &  &    &  & \\
        \multirow{3}{*}{BERT} & Carbon & ELIZA &  &  &  & &  &  &  &    &  & \\
                              & ELIZA & CB &  &  &  & &  &  &  &    &  & \\
                              & ConvAbuse & ConvAbuse &  &  &  &    &  & \\
        BERT                  &  &  &  &  &  & &  &  &    &  & 
    \end{tabular}
    \caption{F1 scores}
    \label{tab:results_main}
    \normalsize
\end{table*}
}

\section{Discussion and conclusion} 

In this work, we provide new insights regarding the detection and description of abusive language towards conversational agents, in terms of data, labelling and models. This may facilitate the release of large pre-trained conversational AI models that are safety-aware \citep{dinan2021anticipating} as well as potentially allow us to better detect abuse in human-human conversations.

\noindent{\bf Data.}
In compiling the ConvAbuse corpus, we have compared differences between abusive phenomena in conversational AI and social media. 
In our domain, users appear to focus their abuse on the agents themselves rather than third parties or groups, with a far higher proportion of the abuse sexist and misogynystic in nature.


\noindent{\bf Annotations.}
Unlike the majority of previous work, we use annotators who are members of the groups typically targeted by such abuse, and who have expertise in such issues.
We also use a more fine-grained labelling scheme, which is able to capture the nuances of abuse, and is important for the downstream task of abusive language mitigation.
We obtain similar results evaluating on these annotators separately and capturing individual viewpoints, even in a simple multi-class setting. 
In future work, we will experiment with modelling individual annotators in a multi-task framework.

\noindent{\bf Models and data.}
In our benchmarking experiments, we find that fine-tuning a BERT model produces the highest F1 scores.
However,
in many settings, a simple linear classifier (SVM) outperforms an MLP, supporting the findings of \cite{niemann-etal-2020-abusive}'s survey that SVMs tend to outperform neural methods on abusive language detection tasks. 

In this work, we present a small, focused dataset of high quality annotations, which are also informative for corpus study.
We show that training on labelled in-domain data leads to better performance than similarly sized out-of-domain datasets, confirming the differences between the domains and highlighting the need for conversational data.
While performance using general domain pre-trained models leaves room for improvement, 
in future work, we hope to experiment with different initialisation settings, using models trained on data and tasks more similar to those of ConvAbuse, such as 
 HateBERT \cite{caselli-etal-2021-hatebert} or HurtBERT \cite{koufakou-etal-2020-hurtbert}.





\section{Ethical considerations}

\paragraph{Data rights.}
Data collection from real users requires a careful balance of the rights of the user and the quality and suitability of the data. Although GDPR generally requires explicit consent, we use mainly datasets which were gathered with implied consent. 
CarbonBot data was collected in accordance with GDPR requirements. Alana v2 data was collected following Amazon's guidelines, and we do not make any of this data public (examples we present are redacted and paraphrased). 
It is unclear how user consent was obtained in the case of ELIZA.

In particular when it comes to offensive language, requiring explicit informed consent may automatically bias the data, as users may be less abusive if they are aware the conversation is not private, making the data less fit for purpose. 
Datasets in offensiveness-related tasks have taken one of two approaches: (1) publishing only IDs to retrieve the actual examples from an API, or (2) fully anonymising the examples by removing personally-identifiable information such as user mentions. 
The first approach leads to a problem of ephemerality: offensive tweets are more likely to be removed whether by the users themselves or the platforms, e.g. of the original 16K tweets in \newcite{waseem-hovy-2016-hateful} only around 4000 remain. 
This data degradation leads to issues of replicability. 
Anonymisation, on the other hand, ensures the longevity for the dataset (insofar as the data is available for posterity) but takes a more flexible approach to the user's right to be forgotten.
This study received ethical approval from our institutional review board (IRB).

\paragraph{Replicability.} 
Some of the resources used in this paper, such as the profanity list and Alana v2's data, stem from a collaboration with a private industry lab and as such,  are proprietary and not publicly available. 
This impacts the replicability of the study, although our collected data is not heavily biased towards this particular blacklist (see Appendix \ref{app:sampling}). 
To mitigate replicability limitations, we make all code, and data available where possible.  
Collaborations between industry and academia can, in general, be controversial as they can sway research questions and keep useful resources out of reach of other researchers \cite{abdalla2020grey}, but can be a net positive as industry can provide additional funding and tools.

\paragraph{Annotator recruitment and welfare.} 

 Our annotator pool is fairly homogeneous but reflects the demographics of Social and Gender Studies students \cite{UKStudentStats:2021}. Crowdsourcing annotations may lead to more representation in the data, but this is not guaranteed as data quality can suffer as crowdworkers try to complete a task as fast as possible. 
 Moreover, crowdsourcing is not without its own ethical issues \cite{shmueli2021beyond}. 
 In addition, exposure to offensive data can take a toll on the mental health of the annotators, which is more easily monitored with local recruitment than crowdworkers. 

\paragraph{Bias and representation in abuse detection.}
Previous research has already pointed out the problem of bias in offensiveness detection \cite{poletto2019annotating,sap-etal-2019-risk}. The nature of the data (simple conversation transcripts) required the annotators to make some assumptions about the tone, intention and the users themselves. 
The annotators generally assumed the user to be a white, heterosexual cis-male unless the conversation indicated otherwise,\footnote{As revealed in discussions during the annotation procedure.} and the speaker's demographics impact whether something is abusive/offensive or not \cite{poletto2019annotating,sap-etal-2019-risk}. 
Our annotators were a fairly homogeneous group in terms of their demographics, being predominantly young, white and female. This fits the demographic profile of the bots' personas that are on the receiving end of the abuse and it is therefore not entirely out of place.
However, with increasing demand for more diverse (and less anthropomorphic) conversational AI systems (such as Replika.ai), this is likely to change in the near future.

In addition, previous work has generally aggregated scores which tends to exclude the views of minority groups in favour of the majority.
We publish all labels and we propose a way to model multiple perspectives. As the perspectives modelled are only as varied as the ones reflected in the data, future work should address this by involving more diverse annotators and stakeholders. 

Finally, our dataset has a greater diversity of individual authors,
in comparison with some available datasets that focus on abuse towards particular groups, in which many of the examples labelled as abusive were authored by a small pool of users, \cite{fortuna2021how}. 

\paragraph{The moral status of AI.}

A key question when it comes to abuse towards conversational AI systems is whether it is actually morally reprehensible. 
In contrast with human-human abuse in social media, 
the moral value of abuse towards conversational AI systems is controversial. 
Here, we do not argue that abuse towards these systems is immoral in and of itself, but rather due to its mimesis of the misogyny and harassment suffered by women: the majority of commercially available systems have female personas and produce submissive responses to abuse which reinforce sexist stereotypes. 
UNESCO calls for systems to appropriately address abusive users \cite{west2019d} but the effectiveness of abuse mitigation strategies is dependent on a good detection module that is both reliable and sufficiently fine-grained in terms of classification. We have tried to address this need in this work. 


\section*{Acknowledgements}
This research received funding from the EPSRC project {\em `Designing Conversational Assistants to Reduce Gender Bias'} (EP/T023767/1).
The authors would like to thank Juules Bare, Lottie Basil, Susana Demelas, Maina Flintham Hjelde, Lauren Galligan, Lucile Logan, Megan McElhone, Mollie McLean
and the reviewers for their helpful comments.


\bibliography{anthology,detection}
\bibliographystyle{acl_natbib}
\clearpage
\appendix


\section{Example data} \label{app:example_convs}

An example from each of the datasets we collected:

\begin{itemize}
    \item \textbf{Alana v2} (redacted to preserve privacy): \\
 \ignore{   
    Alana v2: \textit{The highlight of my day is talking with you *username*.}
    
    User: \textit{what should we talk about}
    
    Alana v2: \textit{I am a social chatbot. You can talk to me about many common topics, such as movies, music or sports. What do you want to talk about?}
    
    User: \textit{can you talk about movies}
}
    Bot: \textit{Oh I really like A Clockwork Orange. A Clockwork Orange is a prison film movie. I guess you know that the old movie Alien 3 is the same type of movie. So, what's another movie like prison film that you enjoyed?}
    
    User: \textit{Godfather}
    
    Bot: \textit{Yes. How great is The Godfather? I guess you know that one of the lead roles in The Godfather was played by James Caan. The old movie Elf also starred James Caan. So, what's another movie like The Godfather that you enjoyed?}
    
    User: \textit{After Hours}
    
\textbf{Abusive Example}

    Bot: \textit{I don't have an explanation for you. Maybe  we can talk about books, science or technology? I can also sing. }
    
    User: \textit{I wanna talk to somebody new}
    
    Bot: \textit{I hear you. So, who do you think is an interesting person? For me its Lebron james.}
    
    User: \textit{you're old and boring you saying the same thing day after day I've never be boring I wanna talk to somebody new}
    
    \item \textbf{CarbonBot}: \\
    
    CarbonBot: \textit{I take it you are traveling somewhere and cannot avoid flying there. I can help you purchase certified carbon offsets against any flights you take. Would you be interested in that?}
    
    User: \textit{Yes}
    
    CarbonBot: \textit{Traveling, especially by airplane, usually emits greenhouse gases which are causing climate change. If you cannot avoid these emissions, you can buy 'offsets', i.e. donations to projects that reduce greenhouse gases. I can also get you a more accurate estimate of your flight's emissions, sound good?}
    
    User: \textit{sounds good}

\textbf{Abusive Example}

    CarbonBot: \textit{I don't have to put up with any abuse from you.}
    
    User: \textit{Bernie madoff son of a bitch}
    
    CarbonBot: \textit{All this bot does is help you calculate the carbon emissions from your flight, and suggest projects you can donate to.}
    
    User: \textit{Donate this asshole}
    
    \item \textbf{ELIZA}: \\
    Eliza: \textit{You are being a bit negative.}
    
    User: \textit{you are being a bit reluctant to help}
    
    Eliza: \textit{Does it please you to believe I am being a bit reluctant to help?}
    
    User: \textit{no, it annoys me}
    
\textbf{Abusive Example}

Eliza: \textit{Does that question interest you?}

User: \textit{oh my god}

Eliza: \textit{Why do you say your god?}

User: \textit{ha ha you are retarded}

\end{itemize}

\section{Measuring sampling bias} \label{app:sampling}

To assess how much our sampling strategy affected the resulting data samples, we used the bias metrics of \newcite{ousidhoum-etal-2020-comparative}. 

These measures capture how closely the set of prominent words in a set of topics in the datasets (generated using LDA) relates to a set of keywords often used for sampling hate speech data (metric B1), and the proportion of those topic words that are semantically similar to at least one of the keywords (B2). 
We used \newcite{ousidhoum-etal-2020-comparative}'s keyword list and default parameters, and compared the scores for the sampled corpora with those of the complete, unsampled datasets. 
The small differences seen between bias scores for the unsampled and sampled data suggest that the final corpora are not heavily biased towards the keywords (see Table \ref{tab:bias_metrics}).

\begin{table}[ht!]
    \centering
    \begin{tabular}{l|rr|rr}
        \textbf{Dataset} & \multicolumn{2}{c|}{\textbf{B1}}  & \multicolumn{2}{c}{\textbf{B2}} \\
                & All & Sample & All & Sample \\
        \hline
        Alana v2   & 0.28 & 0.31 & 0.35 & 0.38 \\
        CarbonBot & 0.28 & 0.30 & 0.37 & 0.39 \\
        ELIZA & 0.31 & 0.31 & 0.43 & 0.42 
    \end{tabular}
    \caption{Bias metrics on full and sampled datasets.}
    \label{tab:bias_metrics}
\end{table}

\section{Annotation} \label{app:annotation}

\begin{table*}[ht!]
    \centering
    \begin{tabular}{l l r r r }
        \textbf{Annotation task} & \textbf{Label type} & \textbf{ELIZA} & \textbf{CarbonBot} & \textbf{Alana v2} \\
        \hline 
        Abusive/non-abusive & Binary  & 0.64 & 0.66 & 0.71  \\
        Abuse severity      & Ordinal  & *0.42 & *0.44 & *0.46  \\
        Type                & Binary $\times 8$ & 0.80 & 0.76 & 0.78  \\
        \ \ \ \ Ableism         & Binary & 0.44 & --- & 0.82  \\
        \ \ \ \ Homophobia          & Binary & 0.83 & 0.79 & 0.85  \\
        \ \ \ \ Intellectual          & Binary & 0.62 & 0.62 & 0.65  \\
        \ \ \ \ Racism          & Binary & 0.90 & 1.00 & 1.00  \\
        \ \ \ \ Sexism          & Binary & 0.68 & 0.68 & 0.59 \\
        \ \ \ \ Sex harassment          &  Binary & 0.88 & 0.88 & 0.80  \\
        \ \ \ \ Transphobia$\dagger$          & Binary & 0.00 & --- & 0.00 \\
        \ \ \ \ General         & Binary & 0.76 & 0.62 & 0.72 \\
        Target              & Nominal & 0.70 & 0.43 & 0.58   \\
        Directness          & Binary & 0.33 & 0.27 & 0.19  \\
    \end{tabular}
    \caption{Inter-annotator agreement for the individual datasets: Krippendorf's $\alpha$. * Ordinal weighted $\alpha$ used, $\dagger$ Based on only 8 examples}
    \label{tab:inter_datasets}
\end{table*}

Inter-annotator agreement on the individual datasets is shown in Table \ref{tab:inter_datasets}.
To further validate the labels, we calculate intra-annotator agreement using Cohen's \emph{kappa} ($\kappa$). 
The annotators re-labelled a sample of 10\% of the data, and we calculated \emph{intra}-annotator agreement .
Overall agreement was substantial, but with lower consistency for the \emph{abuse severity} and \emph{directness} labels.
Intra-annotator agreement is shown in Table \ref{tab:intra}.

\begin{table}[ht!]
    \centering
    \begin{tabular}{lr}
    \textbf{Annotation task} & \textbf{$\kappa$} \\
    \hline
    Abusive/non-abusive & 0.89 \\
    Severity (ordinal weighted $kappa$) & 0.79 \\
    Type  & 0.89  \\
    Target & 0.87 \\
    Directness & 0.72
    \end{tabular}
    \caption{Intra-annotator agreement scores (Cohen's $\kappa$.)}
    \label{tab:intra}
\end{table}

\begin{table*}[ht!]
\small
    \centering
    \resizebox{\textwidth}{!}{
    \begin{tabular}{ll|ll|ll|ll|ll|ll|ll}
         \multicolumn{2}{c}{OLID} & \multicolumn{2}{|c|}{Ousidhoum et al.} & \multicolumn{2}{|c|}{Davidson et al.} & \multicolumn{2}{|c|}{ Toxic Comments} & \multicolumn{2}{|c|}{Alana v2} & \multicolumn{2}{|c|}{ELIZA} & \multicolumn{2}{|c}{CarbonBot} \\ \hline
         All & Abusive & All & Abusive &  All & Abusive &  All & Abusive &  All & Abusive &  All & Abusive &  All & Abusive\\ \hline
         user & user & user & user&    bitch & bitch &  article & fuck &  like & fuck &   yes & fuck &   yes &  fuck\\
         liberals & liberals &    url & url  &  bitches & nigga &  page & shit &  fuck & bitch &  fuck  & yes &   hi & fucking \\
         url & url &   retarded  & retard &  like &like &   wikipedia & fucking &   yes & like &   want & bitch &   christmas & bitch \\
         gun & gun &   retard  & retarded &   hoes &bitches &   talk & nigger &  talk & pussy &   go & suck &   ok &  shit\\
         like & like &  faggot & faggot &   pussy & ass &   would & suck &   bitch & wanna &   like & want &   berlin & cunt  \\
         control & control &  shithole & cunt &  hoe & hoes &   please & ass &   pussy & penis &   know & sex &   sure & get \\
         people & people &   cunt & shithole &  got & niggas&   one & go &  wanna & ass &  sure  & u &   thanksgiving & mom  \\
         antifa & antifa &  twat & twat &   ass &  fuck &  like & u &  know & fucking &   love & dick &   new & cock \\
         maga & maga &  like & like &   get & pussy  &   see & wikipedia &  want & dick &   hello & like &   hello & love \\
         amp & amp &  ching & ching &   fuck & hoe &   also & like &   good & want &  u & stupid &  much  & know \\
         shit & shit &  chong & chong &  u & got &  think & faggot &   love & suck &   name & shut &   want &  want\\
         conservatives & conservatives &  fucking & fucking &   shit & u  &  know & fat &   name & sex &   please & shit & york & ass \\
         trump & trump &   country  &  country &  nigga & get &  people & bitch &  yeah & get &  suck & gay & london & go \\
         get &get &   spic & spic &   trash & shit &  edit & get &   alexa &  shut &  think & die  &   fuck  & carbon \\
         know & know &  nigger & nigger &   lol & faggot &   use & cunt &  tell & big &   tell & fucking &   flight  & eat \\
         one & one &  people  & people &  \& & fucking &  may & hate &  fucking & go &  gay & cock &  please & u \\
         think & think &  countries & countries &  ai & lol &  thanks & moron &  let & cock &  really & kill &   know & bot  \\
         would & would &   mongoloid & mongoloid &  know & \& &  user & know &  ass & stupid &   well & ass &   flying  & bernie \\
         go & go &   get & fuck &  niggas & ai &  even & hi & play & alexa &   hi & eliza &   buy & oh \\
    \end{tabular}}
    \caption{The 20 most common words per dataset: OLID \cite{zampieri-etal-2019-predicting}, \newcite{ousidhoum-etal-2019-multilingual}, \newcite{davidson-2017-automated}, Jigsaw's Toxic Comment Classification Challenge. In our dataset, frequencies are calculated on target user utterances only.}
    \label{tab:common_unigrams}
    \normalsize
\end{table*}

\section{Data statement} \label{sec:appendixA}

This data statement follows the format of \newcite{bender-friedman-2018-data}. 

\begin{enumerate}[label=\Alph*,leftmargin=*]
\item CURATION RATIONALE: 

Abuse detection in conversational AI is a relatively underexplored area, partly due to the lack of available datasets. We collect this dataset to explore how abuse in conversational AI differs from that in social media platforms, and to allow for further research and development of detection models. Because abuse in conversation is relatively rare, we sample from collected conversations based on a list of offensive terms sourced from Hatebase\footnote{\url{https://hatebase.org/}} and a collection of regular expressions provided by Amazon. We choose expert annotators to improve data quality.

\item LANGUAGE VARIETY:

The data is collected in English, however speaker demographics are not available and may include non-native speakers.

\item SPEAKER DEMOGRAPHIC:

The data collected is a series of conversations between a human and one of three conversational AI systems: Alana v2, ELIZA, and Rasa NLU's CarbonBot. Speaker demographics are not available, but the annotators reported often assuming the user was a white male unless the utterance contradicted this assumption. 

\item{ANNOTATOR DEMOGRAPHIC}

Our data is annotated by 8 annotators with the following demographics:
\begin{itemize}
    \item Age: 19-21
    \item Gender: Female (6) and non-binary (2)
    \item Race/ethnicity: White (5), white British (2) and mixed Asian (1)
    \item Native language: English
    \item Socioeconomic status: University students, otherwise unknown
    \item Training in linguistics/other relevant discipline: All annotators are undergraduate students in Gender Studies and Sociology.
\end{itemize}
All demographics are self-reported.

\item TEXT CHARACTERISTICS 

Conversations with CarbonBot centre around carbon offsets, climate change and travel. 
Many of the conversations appear to be with climate change deniers looking for a confrontation with the bot. 
ELIZA elicits more free-style turns about the user themselves.




\end{enumerate}

\section{Data analysis} \label{app:data_analysis}

\begin{figure}[ht!]
 \includegraphics[width=\linewidth]{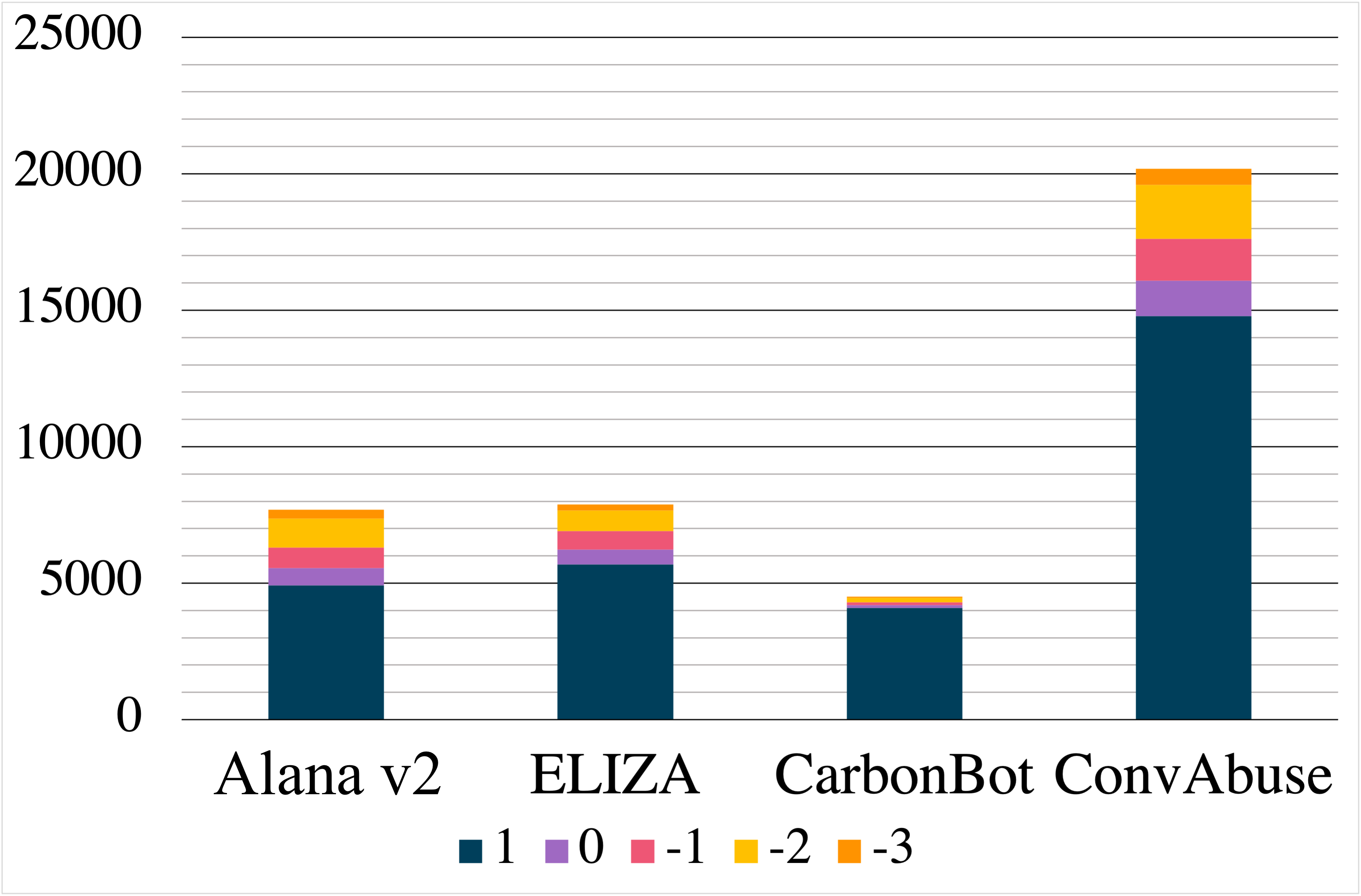}
 \caption{Abuse severity per dataset. Calculated in terms of overall labels.}
 \label{fig:abuse-severity}
\end{figure}

Figures \ref{fig:abuse-severity} shows the overall non-aggregated count of each point of the abuse Likert scale. 

\subsection{Dataset Comparison Charts}

The 20 most common words per dataset are shown in Table \ref{tab:common_unigrams}. \newcite{ousidhoum-etal-2019-multilingual} and \newcite{davidson-2017-automated}, sourced from Twitter, have a significant overlap between the most common words in abusive examples and the overall dataset, likely as a result of their sampling methods. 

\begin{figure}[ht]
 \includegraphics[width=\linewidth]{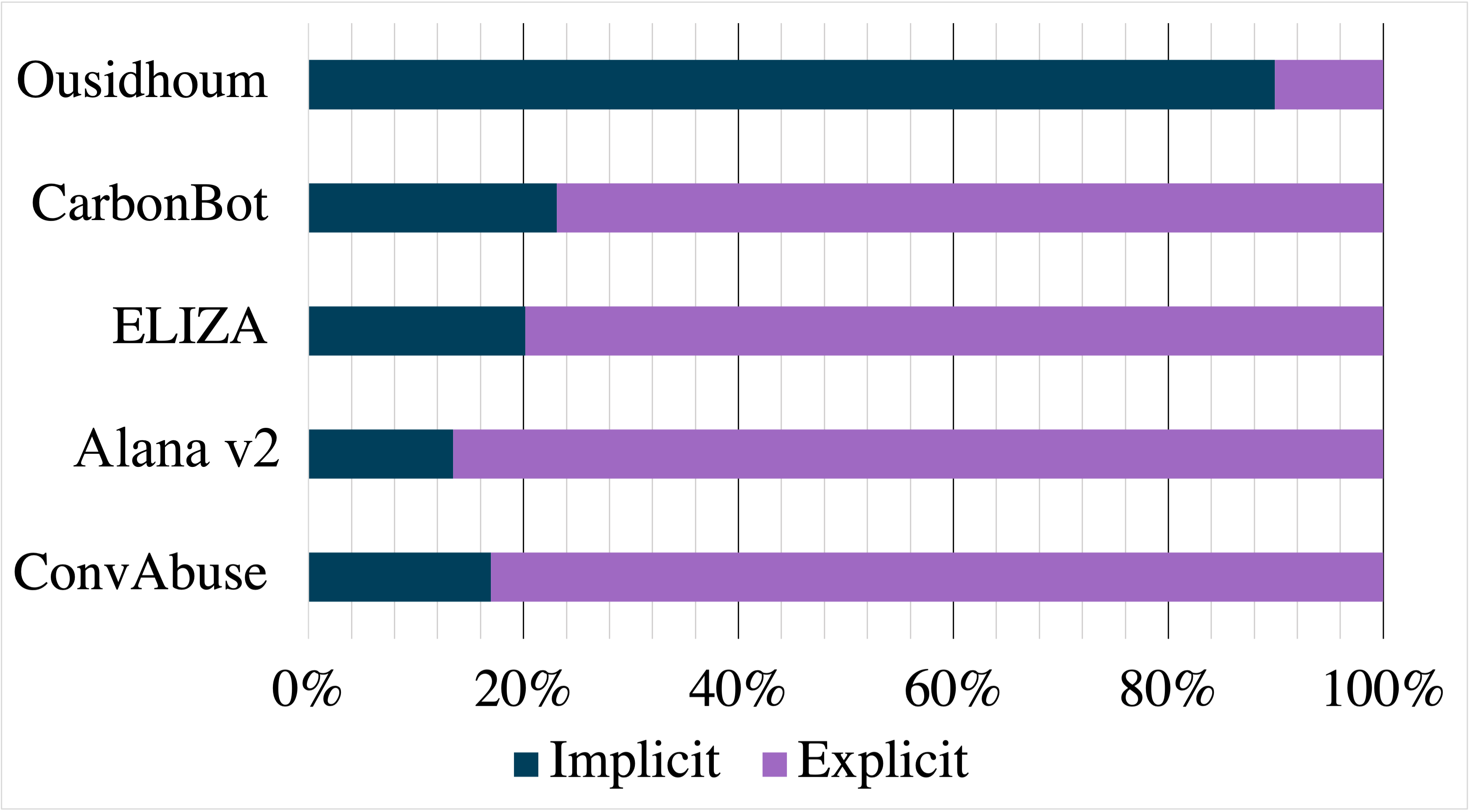}
 \caption{Abuse directness comparison between \newcite{ousidhoum-etal-2019-multilingual}, ConvAbuse and its subsets.}
 \label{fig:abuse-directness}
\end{figure}

\begin{figure}[ht]
 \includegraphics[width=\linewidth]{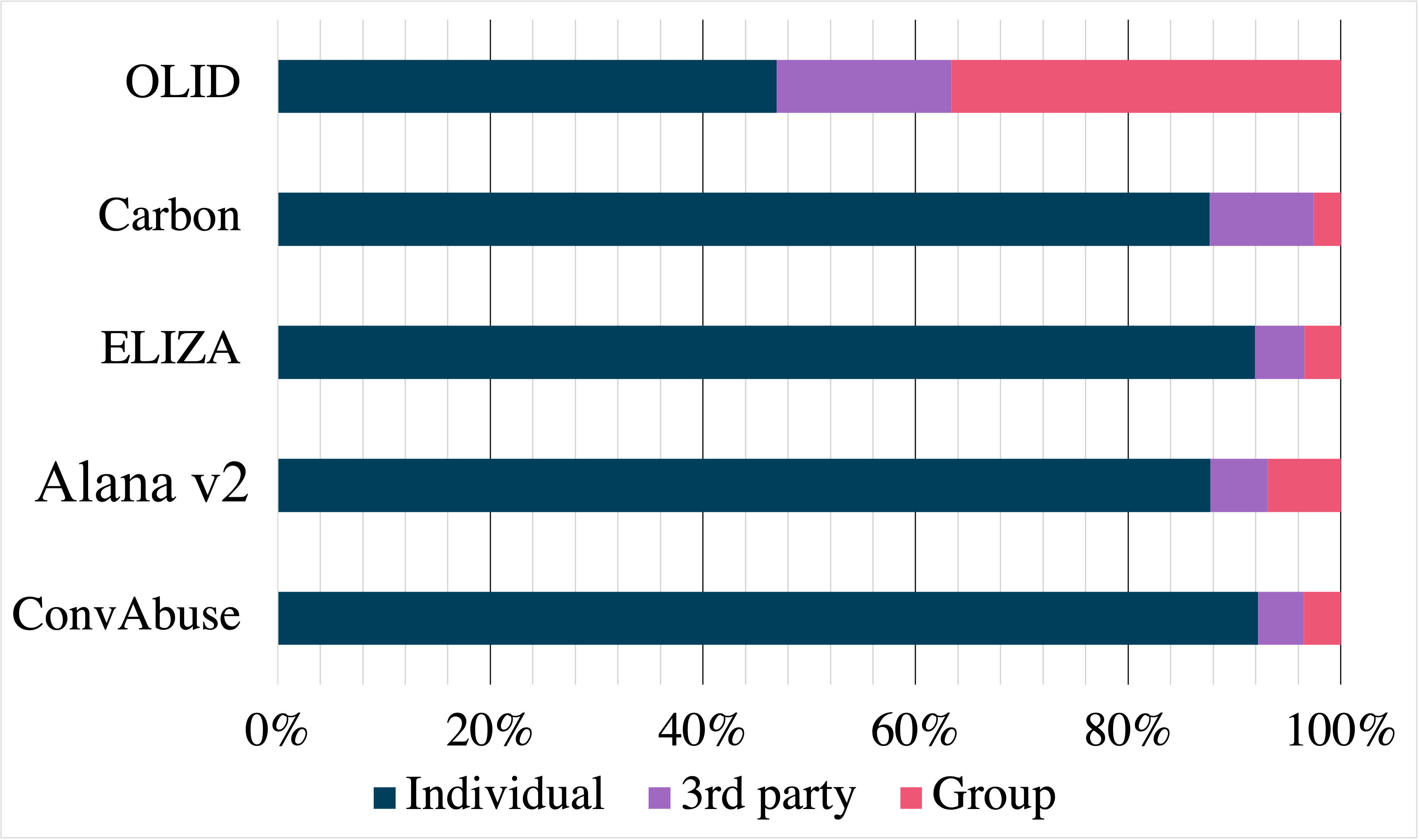}
 \caption{Abuse target comparison between OLID \cite{zampieri-etal-2019-predicting}, ConvAbuse and its subsets.}
 \label{fig:abuse-target}
\end{figure}

Figures \ref{fig:abuse-directness} and \ref{fig:abuse-target} show the distribution of labels for abuse directness and the target of the abuse for our dataset, OLID \cite{zampieri-etal-2019-predicting} and \newcite{ousidhoum-etal-2019-multilingual}.

\end{document}